\documentclass{article}

\usepackage{arxiv}

\usepackage[utf8]{inputenc} 
\usepackage[T1]{fontenc}    
\usepackage{url}            
\usepackage{booktabs}       
\usepackage{amsfonts}       
\usepackage{nicefrac}       
\usepackage{microtype}      
\usepackage{lipsum}
\usepackage{graphicx}
\usepackage{epstopdf}

\title{Efficient long-distance relation extraction with DG-SpanBERT}

\author{
 Jun Chen \\
  King Abduallah University of Science and Technology\\
  Thuwal 23955, Saudi Arabia \\
  \texttt{jun.chen@kaust.edu.sa} \\
   \And
  Robert Hoehndorf \\
  King Abduallah University of Science and Technology\\
  Thuwal 23955, Saudi Arabia \\
  \texttt{robert.hoehndorf@kaust.edu.sa} \\
  \And
  Mohamed Elhoseiny \\
  King Abduallah University of Science and Technology\\
  Thuwal 23955, Saudi Arabia \\
  \texttt{mohamed.elhoseiny@kaust.edu.sa} \\
  \And
   Xiangliang zhang$^\ast$ \\
  King Abduallah University of Science and Technology\\
  Thuwal 23955, Saudi Arabia \\
  \texttt{xiangliang.zhang@kaust.edu.sa} \\
}

\begin{document}
\maketitle
\begin{abstract}
In natural language processing, relation extraction seeks to rationally understand unstructured text. Here, we propose a novel SpanBERT-based graph convolutional network (DG-SpanBERT) that extracts semantic features from a raw sentence using the pre-trained language model SpanBERT and a graph convolutional network to pool latent features. Our DG-SpanBERT model inherits the advantage of SpanBERT on learning rich lexical features from large-scale corpus. It also has the ability to capture long-range relations between entities due to the usage of GCN on dependency tree. The experimental results  show that our model outperforms other existing dependency-based and sequence-based models and achieves a state-of-the-art performance on the TACRED dataset.  
\end{abstract}


\section{Introduction}
Relation extraction aims to discern the semantic relation that exists between two entities within the context of a sentence. For example, in the sentence ``The key was in a chest'', ``key'' is a subject entity and ``chest'' is an object entity. The target for relation extraction is to predict the relation between ``key'' and ``chest'', which is ``Content-Container''. Relation extraction plays a fundamental role in natural language understanding of unstructured text, such as knowledge base population \cite{ji2011knowledge}, question answering \cite{yu2017improved} and information extraction \cite{fader2011identifying}.

The existing solutions for relation extraction can be categorized into dependency-based and sequence-based approaches. Dependency-based models rely on the dependency trees that are able to provide rich structural and syntactic information for classifying relations; see, for example, 
\cite{peng2017cross} and \cite{zhang2018graph}. Sequence-based models directly operate on the word sequences and forgo the information of dependency structures. For example, the model described in \cite{wang2016relation} relies on a multi-level attention mechanism to capture the attentions regarding target entities and relations. Bidirectional Long Short-Term Memory (LSTM) is applied on sentences   
to capture the semantic features in \cite{zhou2016attention}. Recently, BERT-related models \cite{joshi2019spanbert,wu2019enriching,soares2019matching} have shown their ability to improve relation extraction tasks and achieve state-of-the-art results.  

Although BERT-based models are strong on learning rich semantic features, 
they may not effectively capture the long-range syntactic relations.
For example, in the sentence ``Arcandor said in documents filed Wednesday with a district court in Essen, where it is based, that the 15 companies include Corporate Service Group GmbH'', with  ``Arcandor'' as the subject and ``Corporate Service Grroup GmbH'' as the object, it is difficult for sequence-based models to extract features between such long-distance entities. 
Therefore,  we propose DG-SpanBERT model, which is the first to combine BERT-related model with Graph Convolutional Network (GCN) on relation extraction. Specifically, our model groups BERT sentence-embedding in a dependency tree structure and then uses a GCN network to extract features from the tree. DG-SpanBERT thus has a unique advantage that leverages the local features and better captures long-distance relations than other models. Our model obtains an F1-score of 71.5\% in TACRED and achieves a state-of-the-art performance.

\section{Related Work}

Early research efforts on relation extraction were based on statistical methods. \cite{Zelenko:2003:KMR:944919.944964} introduced a shallow tree-based kernel to determine the relation between two entities. \cite{Bunescu:2005:SPD:1220575.1220666} parsed the sentence into a dependency graph and recognized the relation types from the shortest path between two named entities. \cite{mcdonald2005simple} predicted the relations by finding maximal cliques of entities. \cite{mintz2009distant} dealed with a long sentence by adding syntax features to a statistical classifier. 
\
Neural-based models have achieved a lot of success in relation extraction. \cite{zeng2014relation} used a deep convolutional neural network (CNN) on the relation classification, which outperformed the previous statistical machine learning methods. Then many neural models followed, such as attention-based Bidirectional LSTM \cite{zhou2016attention} and CNN with two levels of attention \cite{wang2016relation}.  
With the born of BERT \cite{devlin_bert}, solutions to relation extraction reached a new high-level \cite{shi2019simple}. 
The most significant improvement was achieved by SpanBERT \cite{joshi2019spanbert},

which masks continuous spans of text instead of random tokens in pre-training that results in better word embedding and outperforms the BERT model for many NLP tasks, in particular for relation extraction. 
SpanBERT is the most recent 
state-of-the-art model 
for relation extraction  across the family of different methods. 
\
Another stream of solutions demonstrates that incorporating a dependency tree into neural models could improve relation extraction. \cite{liu2015dependency} designed a dependency-based neural network (DepNN) that uses the recursive neural network to model the sub-tree and also incorporates the CNN to learn features from the shortest path between two entities. \cite{xu2015classifying} argued that a dependency tree could help a neural model to capture the long-distance relations 
and thus, they applied LSTM to the shortest dependency path between two target entities. More recently, \cite{zhang2018graph} proposed to use GCN
on a pruned dependency tree that is tailored to relation extraction. 

Our work is inspired by the superior performance of the SpanBERT model \cite{joshi2019spanbert} and the capability of a dependency tree on grouping long clauses together and extracting the grammar structure of the sentence \cite{zhang2018graph}. We show that our DG-SpanBERT model achieves the best performance on the TACRED dataset.

\section{DG-SpanBERT model}
In this section, we first describe the SpanBERT model, and then explain how we use GCN 
to learn  features of the dependency tree. Finally, we introduce our model architecture that combines SpanBERT and GCNs for relation extraction.

\subsection{Pre-trained Model SpanBERT}

The SpanBERT model \cite{joshi2019spanbert} extends BERT model \cite{devlin_bert} 
to better represent and predict spans of text. SpanBERT differs from BERT in three ways: First, during pre-training, spans of tokens are masked, rather than  individual tokens. Second, in SpanBERT, a span boundary objective (SBO) is introduced and only the tokens at the span's boundary are predicted, which represents the entire masked span. Third, SpanBERT only trains single contiguous segments of text for the masked language modeling (MLM) task. With these modifications of the original BERT, the SpanBERT model shows a significant improvement for many NLP tasks. Usually, when we input a sequence of words or sub-word tokens $X = (x_1,..., x_n)$ into a BERT-like model, it can produce a contextualized vector representation for each token: $O = vec (x_1,...,x_n)$. The BERT model also appends the token '[CLS]' to the beginning of the sequence $X$, and the first token in the final output hidden states is used to represent the whole sentence embedding for classification tasks.

\subsection{Graph Convolutional Network}

The GCN model \cite{kipf2016semi} has been popularly used for learning  graph representation. 
Given a graph with $n$ nodes, we can use an $n \times n$ adjacency matrix $A$ to represent the graph structure, where $A_{ij} = 1$ if there is an edge connecting between node $i$ and node $j$, $A_{ij} = 0$ otherwise.  
For a node $i$, its representation  at the $l$-th layer is obtained by 
\begin{equation} h_i^l = \sigma(\sum_{j=1}^{n} A_{ij} W^{l} h_{j}^{l-1} + b^{l})
\end{equation}
where $h^{l-1}$ is the node representation from $(l-1)$-th layer, $W^{l}$ is the weight, 
$\sigma$ is the nonlinear activation function and $b^{l}$ is a bias term. This operation updates the representation of node $i$ by aggregating its neighborhood  via a convolution kernel. 
\subsection{Our proposed  DG-SpanBERT model}

\begin{figure}[t]
\centering
    \includegraphics[width=0.7\linewidth]{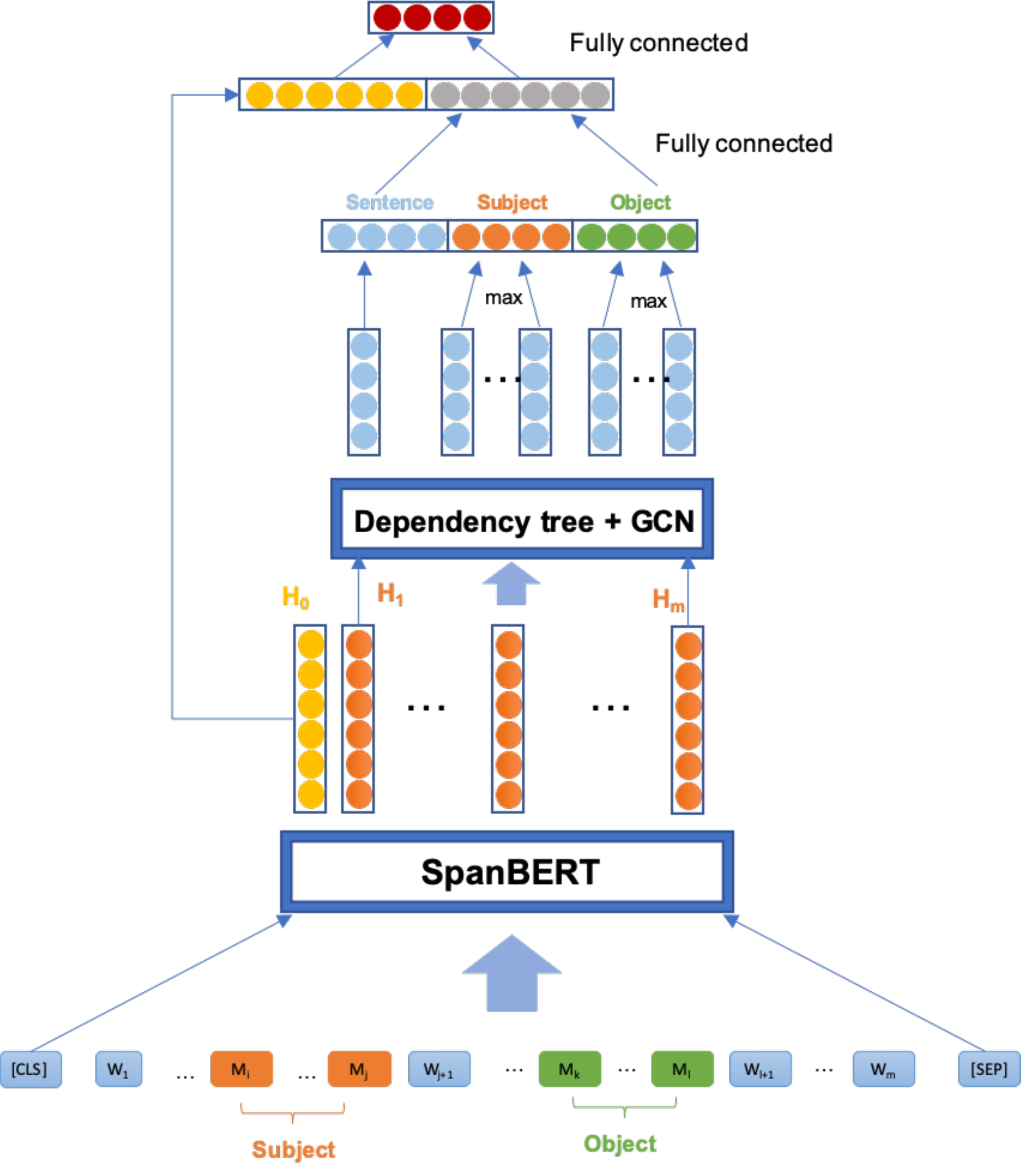}
\caption{Architecture of the DG-SpanBERT model}
\label{fig:figure1}
\end{figure}

Given a sentence $X =[x_1,...,x_n]$ and two non-overlapping entity spans, $X_s = [x_{s_1},...,x_{s_n}]$ and $X_o = [x_{o_1},...,x_{o_n}]$, where $x_i$ is the $i^{th}$ token, and $X_s$ and $X_o$ correspond to the subject entity and object entity, respectively. A relation extraction task, by definition, is to classify the relation $r \in \mathbb{R}$ (predefined relations) between $X_s$ and $X_o$. 
\
Figure \ref{fig:figure1} shows the overall architecture of our model. To enable the SpanBERT module to capture the location of $X_s$ and $X_o$ without overfitting to  these two token spans, we replace each subject and object token with "[unused\_index]", where the index is defined corresponding to the token name entity recognition (NER) types. For example, after we replace subject and object entities with special tokens, the sentence \emph{Alan Gross was working in Cuba for a development contractor when he was arrested in December} with subject \emph{Alan Gross} and object \emph{Cuba} is converted to:\\
``[CLS] [unused\_1] [unused\_1] \emph{was working in} [unused\_2] \emph{for a development contractor when he was arrested in December} [SEP]''.
\
\
\
\
\
SpanBERT tokenizes each word into its sub-word representations. The natural language dependency parser, such as the Stanford Parser \cite{socher2013parsing}, parses the sentence into a dependency tree. A dependency parser usually groups phrases together and analyzes the grammatical structure of a sentence. Due to the semantic gap between different tokenization systems of SpanBERT and dependency parser, we align the SpanBERT tokens with dependency parser tokens via random sampling. According to our random sampling strategy, if one word is tokenized into several sub-words by SpanBERT, we  randomly select one sub-word to represent the whole word. This operation has an advantage that each sub-word can be mapped to its original word, based on which we build a more meaningful and smaller dependency tree, comparing to that using all subwords.  

Suppose the final hidden state output from SpanBERT is $H = [h_0,...,h_n]$, where the first token $h_0$ represents the whole sentence embedding. The rest of the hidden states $[h_1,...,h_n]$ will be used as the initial representation of  each token in $X:[x_1,...,x_n]$. We then convert the sentence into a dependency tree structure, where a node is a token with an attributed vector $h_i$. To allow the information in layer ${l-1}$ to be properly carried to the next layer ${l}$ in the graph convolution operation, we add a self-loop to each node for forming a graph, represented by  an adjacency matrix $A$. As shown in Eq. (1), GCN iteratively updates the node representation $h_i^l$ layer by layer, through integrating the node's neighborhood information. Our GCN operates on  $H$ from SpanBERT and $A$ of dependency tree, and thus results in a seamless integration of rich lexical features and semantic relevance of entities in a long range. That's the key to make our DG-SpanBERT perform better than other models on extracting long-distance relations.

After applying an $L$-layer GCN over the adjacency matrix $A$ and SpanBERT hidden states $H$, we obtain the convolved features for each token. To generate a sentence representation, we use  max pooling  to map from $n$ token vectors to one vector, 
\begin{equation} h_{sentence} = f(GCN(H,A))
\end{equation}
where $f$ is the pooling function: $\mathbb{R}^{d \times n} \rightarrow \mathbb{R}^{d}$. The same max pooling is applied to extract subject and object entity representation $h_s$ and $h_o$, respectively.



The representation $h_{sentence}$, $h_{s}$ and $h_{o}$
are then concatenated  and fed into a feed-forward neural network (FFNN), which outputs a transformed feature representation. Finally, we concatenate the SpanBERT sentence representation $h_0$ and the transformed GCN feature representation to produce the final representation, as shown in the last layer of Figure \ref{fig:figure1}:
\begin{equation} h_{final} = [h_{0}; W_{c}([h_{sentence}; h_{s}; h_{o}])+b_c]
\end{equation}
\
In the end,  $h_{final}$ is fed to another FFNN with softmax operation to output a probability distribution over relations.
The loss function is defined based on cross entropy for the end-to-end model parameter optimisation. 
\
\section{Experimental Evaluation}
\subsection{Dataset and Experimental Setup}
\
\
The TACRED dataset \cite{zhang2017position} contains over 106K sentences, and it covers 41 relation types (e.g., per:schools\_attended and org:members) and one ``no relation'' label to describe the relation between the subject and the object in the sentences. The types of subject and object mentions are categorized into 17 fine-grained types, including person, organization and location, etc. We replace the subject and object entities with their NER tags by following the entity masking schema as described in \cite{zhang2017position}.

We train our SpanBERT model and the GCN part with a different learning rate (3e-5 and 0.01, respectively). We follow the SpanBERT settings for other hyperparameters.  To make an appropriate comparison, all evaluated models in this paper are trained with the same setting. The GCN hidden states dimension is set to be 400.

\begin{table}[ht]
	\centering
	\begin{tabular}{lccc}
	\hline
    \textbf{Model} & \textbf{P} & \textbf{R} & \textbf{F1}   \\
	\hline
	C-GCN & 68.9& 66.3& 67.6\\
	C-AGGCN & 69.6& 66 & 67.8\\
	\hline
	Google Bert(base) & 68.1 & 67.7 & 67.9 \\
    SpanBert(base) & 67.6  & 68.6 & 68.1\\
    DG-SpanBERT(base) & 68.3 & \textbf{72.1} & 70.2\\
    \hline
    Google Bert(large) &69.2 & 72 & 70.6\\
    SpanBert(large) &71.2& 70.4 & 70.8\\
    
    DG-SpanBERT(large) &\textbf{71.4} &71.6 &\textbf{71.5}\\
   
    \hline
	  \end{tabular}
	\caption{Compare DG-SpanBERT with other models on TACRED dataset.  Bold indicates the best performance among all. } 
    \label{Tab:01}
		 \end{table}
\subsection{Results and Discussion}
We compare our model with previously existing dependency-based and sequence-based neural models that do not introduce external features. 
The results in Table \ref{Tab:01} show that   DG-SpanBERT  outperforms all the sequence-based models by at least 0.8 F1 and outperforms all the dependency-based models by at least 3.7 F1. That's saying: our model achieves a new state-of-the-art performance on  TACRED. Moreover, if we only consider the base pre-trained models, our model outperforms other BERT-related models by at least 2.1 F1. 

\begin{figure}[ht]

\centering
    \includegraphics[width=0.5\linewidth]{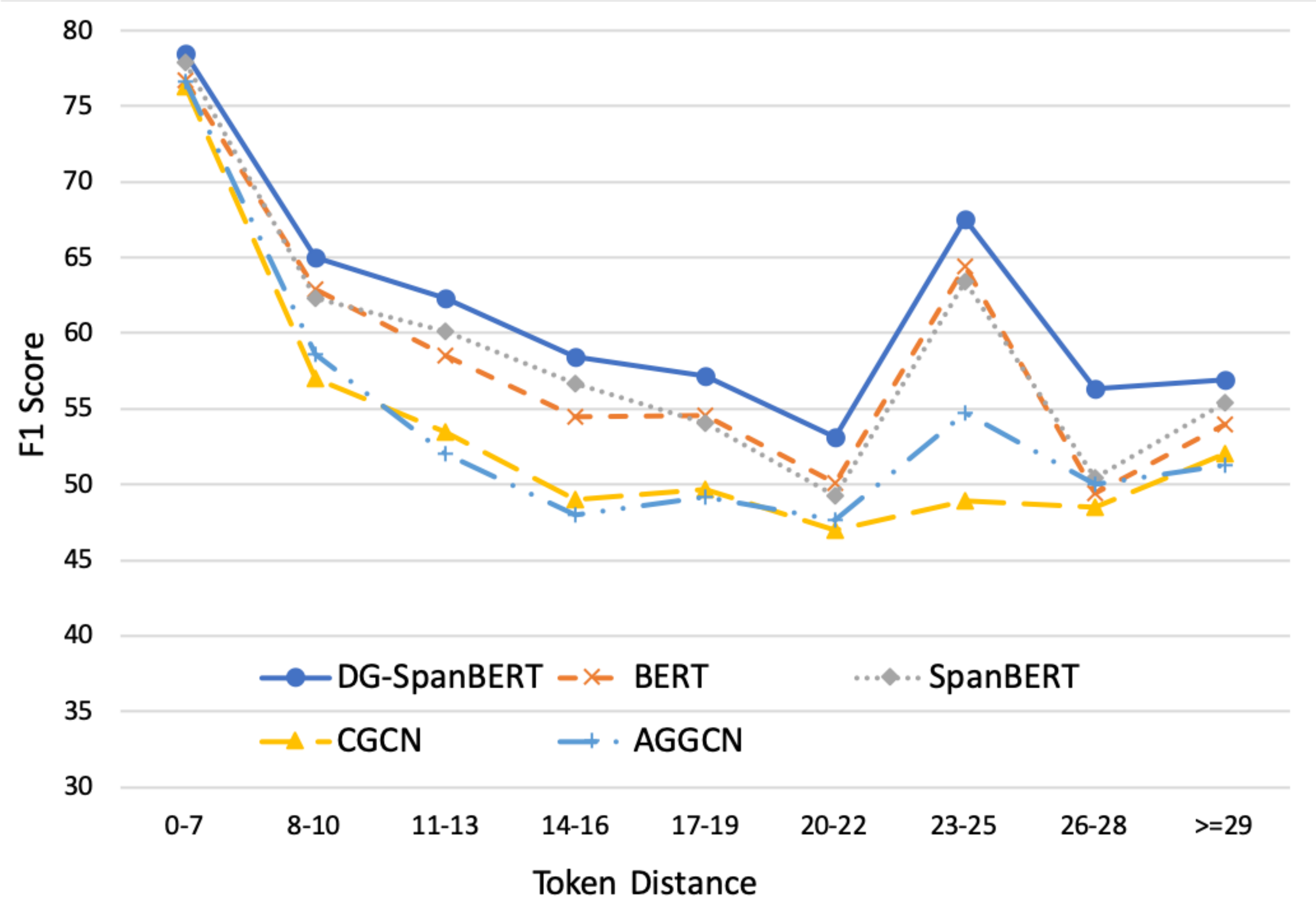}
\caption{Comparison of BERT, SpanBERT, DG-SpanBERT, CGCN and AGGCN w.r.t. the distance between the subject and object.}
\label{fig:figure2}
\end{figure}
Due to the natural integration of $H$ (from SpanBERT) and $A$ (from dependency tree) by GCN, our DG-SpanBERT is expected to perform well on predicting long-distance relations. Therefore, we evaluate the performance of several models on predicting relations with different number of tokens between the subject and object. As shown in Figure \ref{fig:figure2}, all models perform similarly on relations with distance less than 8, since it is an easy task to predict short-distance relations. However, for long-distance relations ($\ge$8), our DG-SpanBERT significantly outperforms all other baselines.

\begin{table}[ht]
	\centering
	\begin{tabular}{lcc}
	\hline
     \textbf{Model} &  \textbf{Avg F1}   \\
	\hline
	CGCN & 49.8\\
	AGGCN & 50.4\\
	Google BERT & 55.1  \\
	SpanBERT & 55.6  \\
	DG-SpanBERT & \textbf{58.7} \\
	\hline

  \end{tabular}\\
\caption{Compare average F1 performance for those token distances $\geq$ 11 tokens between the subject and object. Bold indicates the best performance among all.}
\label{Tab:02}
	 \end{table}
Since the average distance between the subject and object is approximately 12 tokens, we report the average F1 performance of these models on predicting relations   
with span distance $\geq$ 11 tokens,  in  Table \ref{Tab:02}. We observe that 
DG-SpanBERT outperforms other BERT-related models by at least 3.1 F1, 
and  other dependency-based models by at least 8.3 F1.  
This significant improvement confirms the effectiveness of our DG-SpanBERT model.



\section{Conclusion}
In this paper, we proposed a novel neural network model based on SpanBERT and a graph convolutional network for relation extraction. We showed how our model achieves the state-of-the-art results on a large-scale TACRED dataset. We also showed how our model is particularly effective at capturing long-distance relations compared to other models.

\bibliographystyle{unsrt}  
\bibliography{references}  


\end{document}